\documentclass[lettersize,journal]{IEEEtran}
\usepackage{amsmath,amsfonts}
\usepackage{algorithmic}
\usepackage{array}
\usepackage[caption=false,font=normalsize,labelfont=sf,textfont=sf]{subfig}
\usepackage{textcomp}
\usepackage{stfloats}
\usepackage{url}
\usepackage{verbatim}
\usepackage{graphicx}
\usepackage{mwe}
\usepackage{cuted}
\usepackage{booktabs}
\usepackage{multirow}
\usepackage{makecell}
\usepackage{xcolor}
\usepackage{colortbl}

\definecolor{bestcolor}{HTML}{FF9999}   
\definecolor{secondcolor}{HTML}{FFCC99} 

\def\BibTeX{{\rm B\kern-.05em{\sc i\kern-.025em b}\kern-.08em
    T\kern-.1667em\lower.7ex\hbox{E}\kern-.125emX}}
\usepackage{balance}
\begin{document}
\title{URL-Sim: Unified and Regulated LiDAR Simulation for High-Fidelity Reconstruction of Dynamic Driving Scenes}
\author{Kefei Qian, Xipeng Feng, Jie Li, Long Cao, Yuhan Chen, Xiangfei Huang, Guofa Li
\thanks{\textbf{Preprint Notice: This work has been submitted to the IEEE for possible publication. Copyright may be transferred without notice, after whichthis version may no longer be accessible.}

This work was supported by the National Natural Science Foundation of China under Grant 52272421. (\textit{Corresponding author: Guofa Li})

All authors are with the College of Mechanical and Vehicle Engineering, Chongqing University, Chongqing, 401331, China. (email: qiankf@stu.cqu.edu.cn; 19515669228@163.com; jieli@cqu.edu.cn; caolong@stu.cqu.edu.cn; 20240701028@stu.cqu.edu.cn; hxf0213@stu.cqu.edu.cn; liguofa@cqu.edu.cn)}}

\markboth{}%
{}

\maketitle

\begin{abstract}
Reconstructing high-fidelity LiDAR simulation scenes from real-world driving logs is essential for cost-effective safety validation of autonomous driving systems. While simulators based on Neural Radiance Fields (NeRF) achieve realistic scene reconstruction, their large-scale adoption is limited by slow rendering speeds. Recent 3D Gaussian splatting (3DGS) approaches have improved rendering efficiency through explicit primitive representation and differentiable rasterization. However, existing 3DGS-based methods remain largely restricted to rigid vehicles and produce redundant primitives through unconstrained densification. To address these limitations, we propose URL-Sim, a unified and regulated dynamic LiDAR simulation framework. URL-Sim constructs a dynamic Gaussian scene graph that decouples the modeling of traffic participants, including vehicles, pedestrians, and cyclists, enabling accurate reconstruction of heterogeneous dynamic objects within a single framework. It further introduces a regulated Gaussian densification strategy that ranks primitives with multi-dimensional importance scores and allocates Gaussians to the most critical regions while keeping the total primitive count bounded. Experiments on the Waymo and KITTI-360 datasets demonstrate that URL-Sim achieves higher quantitative accuracy and visual quality than existing baselines under comparable primitive budgets, and accurately recovers the geometry and poses of diverse dynamic traffic participants.

\end{abstract}


\section{Introduction}
\IEEEPARstart{L}{iDAR} sensors are integral to environmental perception in advanced autonomous driving systems \cite{1}, making the driving-scene data they provide critical for the testing and validation of perception and planning algorithms \cite{2}. Since traditional on-road data collection is inefficient and offers limited scene coverage, there is an increasing need for efficient and realistic LiDAR simulation methods \cite{3}.

Traditional LiDAR simulation, whether physics-engine-based virtual worlds \cite{CARLA,AirSim,6} or explicit geometric reconstruction \cite{LiDARsim,8,9,PCGen}, suffers from substantial sim-to-real gaps, cumbersome pipelines, and limited support for dynamic scenes. Neural Radiance Fields (NeRF) \cite{nerf} narrow the sim-to-real gap for LiDAR simulation using continuous implicit representations and differentiable volume rendering \cite{NFL,LiDAR-NeRF,UniSim,NeuRAD}. Some methods incorporate spatio-temporal constraints \cite{LiDAR4D} and bounding box tracking \cite{DyNFL} to handle dynamic scenes. However, the dense ray sampling inherent in volume rendering causes substantial computational overhead. The consequent slow training and rendering speeds cannot meet the requirements of large-scale modeling and real-time closed-loop simulation.

Recent 3D Gaussian splatting (3DGS) \cite{3dgs} has emerged as an effective alternative to NeRF, achieving order-of-magnitude improvements in rendering efficiency through its explicit Gaussian representation and hardware-accelerated rendering. While current mainstream methods \cite{4dgs,drivinggs,omnire,streetgs} primarily focus on camera data synthesis, seminal works such as GS-LiDAR \cite{gslidar} and LiDAR-GS \cite{lidargs} have successfully adapted 3DGS to the LiDAR imaging model by designing differentiable splatting pipelines tailored to LiDAR's active perception mechanism. To handle dynamic scenes, LiDAR-RT \cite{LiDAR-RT} and SplatAD \cite{SplatAD} employ a dynamic-static separation strategy to model rigid vehicles separately. Despite these advances, existing Gaussian-based LiDAR simulation methods face two primary challenges. First, current dynamic scene simulations primarily focus on rigid objects and provide limited support for non-rigid entities such as pedestrians and cyclists, who are vulnerable road users central to high-risk edge cases. Second, existing methods incur substantial computational overhead, driven by the inherent sparsity and non-uniform angular distribution of LiDAR point clouds, combined with the unrestricted densification of the Gaussian paradigm. These issues reduce training efficiency and simulation performance, potentially triggering training failures and constraining practical scalability and scene adaptability.

To address these challenges, URL-Sim is proposed as a unified and regulated LiDAR simulation framework for the high-fidelity reconstruction and controllable editing of dynamic urban driving scenes. URL-Sim constructs a dynamic Gaussian scene graph that decouples individual dynamic entities from the static background for independent modeling. Rigid objects such as vehicles are simulated via standard rigid transformations. For non-rigid entities such as pedestrians, the Skinned Multi-Person Linear (SMPL) \cite{smpl} model is introduced to serve as a deformation prior. Specifically, the Gaussian primitives of each pedestrian are bound to the SMPL mesh vertices, and the pose and shape parameters of the model are utilized to drive the skinning deformation of the Gaussians. For template-free non-rigid objects such as cyclists, a lightweight deformation network is employed to directly learn and optimize the local deformation parameters. Together, these three representations enable unified modeling and simulation of the principal traffic participants in autonomous driving.

To curb the unconstrained expansion of primitives and the redundant accumulation of Gaussians inherent in the original Adaptive Density Control (ADC), a regulated Gaussian densification (RGD) strategy is proposed. RGD imposes a predefined global upper bound on the total Gaussian count and allocates a fixed budget for each densification step following a quadratic growth curve, establishing a deterministic training schedule. A Gaussian importance scoring system is constructed by integrating spatial gradients, image-domain features, and rendering accumulation attributes. Guided by these scores, targeted densification and adaptive pruning are executed within the allocated single-round budget. This controllable mechanism limits redundant proliferation and achieves higher rendering performance under a strictly bounded primitive count.

In summary, our main contributions are as follows:
\begin{itemize}
\item URL-Sim is presented as a unified and regulated LiDAR simulation framework that models diverse dynamic non-rigid objects, addressing a relatively under-explored problem in LiDAR simulation.

\item A regulated Gaussian densification strategy is proposed to balance simulation accuracy and computational overhead through quadratic budget allocation and multi-dimensional scoring.

\item Extensive experiments on Waymo and KITTI-360 validate accurate recovery of non-rigid actor geometry and poses, as well as competitive 2D perceptual quality and 3D geometric accuracy.

\end{itemize}

\section{Related work}
\subsection{LiDAR Simulation and Novel View Synthesis}

Classical LiDAR simulation relies on virtual platforms such as CARLA \cite{CARLA} and AirSim \cite{AirSim} for environmental control and closed-loop interaction \cite{6}, but inherently suffers from severe sim-to-real domain gaps. Efforts to bridge this gap through physical modeling \cite{LiDARsim} and real-world asset libraries \cite{8,9} are hindered by cumbersome procedures and high labor costs.

Neural Radiance Fields (NeRF) \cite{nerf} have advanced LiDAR simulation through implicit spatial modeling and volume rendering, narrowing the sim-to-real gap. Early frameworks including NFL \cite{NFL} and LiDAR-NeRF \cite{LiDAR-NeRF} integrate the LiDAR imaging process into NeRF to enable high-quality novel view synthesis. For complex dynamic scenes, UniSim \cite{UniSim} and NeuRAD \cite{NeuRAD} utilize feature grids and precise physical modeling to reconstruct and recompose traffic participants. DyNFL \cite{DyNFL} and LiDAR4D \cite{LiDAR4D} further isolate dynamic objects and capture temporal correlations to achieve editable digital twins and high-fidelity 4D spatio-temporal synthesis. Nevertheless, the reliance on high-frequency dense sampling induces considerable computational overhead and slow rendering speeds, precluding real-time closed-loop deployment.

3D Gaussian splatting (3DGS) \cite{3dgs} offers a real-time alternative via its explicit primitive representation and highly parallelized rasterization. Early 3DGS-based methods such as GS-LiDAR \cite{gslidar} and LiDAR-GS \cite{lidargs} render substantially faster than NeRF-based counterparts with comparable reconstruction quality, but are limited to static scenes and lack editability. LiDAR-RT \cite{LiDAR-RT} separates dynamic and static components using bounding boxes and tracking IDs to enable customizable scene editing. SplatAD \cite{SplatAD} and SIMULI \cite{SimULi} similarly rely on 3DGS for joint camera-LiDAR simulation. However, these approaches are predominantly restricted to rigid vehicles and overlook non-rigid objects such as pedestrians and cyclists. Moreover, current frameworks lack resource-efficient mechanisms for controlling redundant Gaussians resulting from unconstrained densification.

\subsection{Gaussian Resource Optimization}

Although 3DGS demonstrates promising sensor simulation capabilities, its heuristic cloning and splitting generate redundant Gaussian primitives, inflating VRAM consumption and storage \cite{steepest,apn}. Various approaches have been proposed to optimize the computational efficiency of 3DGS for large-scale deployment.

The first category targets post-processing of redundant models via compression and pruning. Compact 3DGS \cite{compact3dgs} and Compressed 3DGS \cite{compressed3dgs} employ vector quantization and entropy coding to compress Gaussian parameters, reducing storage without altering underlying spatial geometry. Pruning strategies eliminate Gaussian primitives with marginal visual contributions, guided by metrics such as multi-view consistency, uncertainty scores, and global importance \cite{reducing,pup,lightgaussian}. Although model volume is substantially reduced, excessive pruning at extreme compression ratios frequently leads to severe visual degradation and loss of foreground geometric details.

The second category optimizes densification during training to prevent redundant generation at its source, replacing unconstrained growth with rigorous control. SteepGS \cite{steepest} and Taming 3DGS \cite{taming} minimize redundant sub-Gaussians through analytical optimization and fixed-budget constructive pipelines. BUGS \cite{bus} employs a region-consistency-based bi-directional growing mechanism to adaptively regulate Gaussian density for compact scene reconstruction. For LiDAR geometry, GTLR-GS \cite{gtlr} and GeomGS \cite{geomgs} incorporate curvature-adaptive refinement and geometric confidence scores to govern densification. These works demonstrate that proactive densification is critical for compact, high-fidelity reconstruction.

Nevertheless, these strategies are designed for visual modalities and rely on dense pixel-level observations. When applied directly to LiDAR simulation, they may misallocate Gaussian primitives by overlooking sparse ray-level geometry and intensity cues, thereby compromising range accuracy and foreground structural fidelity.

\section{Methodology}
\begin{figure*}[t]
    \centering
    \includegraphics[width=\textwidth]{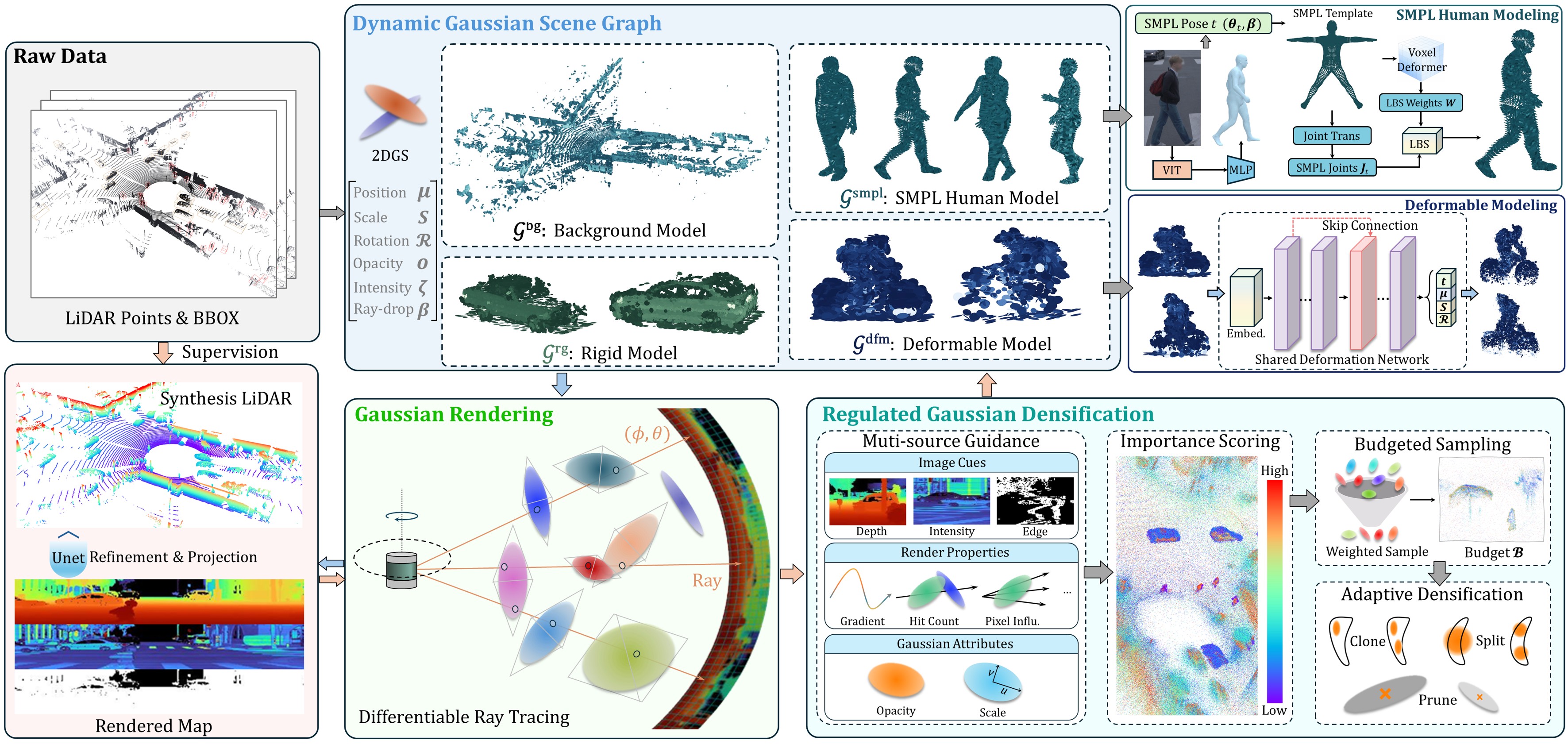}
    \caption{Pipeline of URL-Sim. Raw data is parsed into a Dynamic Gaussian Scene Graph with specialized representations for static backgrounds and dynamic actors (rigid, SMPL human, deformable). These are processed through differentiable LiDAR ray tracing \cite{LiDAR-RT}, while the RGD strategy concurrently optimizes Gaussian distribution via multi-source importance scoring and budgeted sampling. A post-processing U-Net refines the final LiDAR output.}
    \label{fig:overview}
\end{figure*}

The overall pipeline of URL-Sim is illustrated in Fig. \ref{fig:overview}. Given raw LiDAR point clouds and object bounding boxes, URL-Sim builds a dynamic Gaussian scene graph that models the static background and dynamic instances separately. Each instance is transformed to world coordinates at its timestamp and composed with the background. High-fidelity sensor data is synthesized through a Gaussian rendering pipeline based on differentiable ray tracing. Alongside this rendering pipeline, a
regulated Gaussian densification (RGD) strategy is introduced during the training phase to mitigate the resource redundancy caused by unconstrained primitive growth.

\subsection{Dynamic Gaussian Scene Graph}
URL-Sim decomposes the raw LiDAR scene into a static background model $\mathcal{G}^{bg}$ and various dynamic instance models, comprising rigid objects $\mathcal{G}^{rg}$, SMPL-based humans $\mathcal{G}^{smpl}$, and other non-rigid deformable entities $\mathcal{G}^{dfm}$. Each model is parameterized by a distinct set of Gaussian primitives, each of which independently learns its specific geometric and reflectance properties. To represent scene geometry more accurately, the 2D Gaussian splatting (2DGS)\cite{2dgs} is selected as the basic element, which shares a common set of parameters:

\begin{equation}
\label{deqn_ex1}
\mathcal{G}=\{\mu,\mathcal{S},\mathcal{R},o,\zeta,\beta\}
\end{equation}
\noindent Specifically, each Gaussian primitive is parameterized by a mean position $\mu$, scaling factor $S$, rotation quaternion $R$, opacity $o$, intensity $\zeta$, and ray drop probability $\beta$, all of which are learnable. The variables $S$ and $R$ jointly define the covariance matrix as $\mathbf{\Sigma}=RSS^{\top}R^{\top}$. Notably, $\zeta\in\mathbb{R}^1$ and $\beta\in[0,1]$ are LiDAR-specific physical attributes representing the reflectance and no-return characteristics of real-world laser beams, respectively. Following the formulation in LiDAR-RT \cite{LiDAR-RT}, the ray drop probability $\beta$ is derived via the Softmax normalization of two logit parameters, $\beta_{drop}$ and $\beta_{hit}$. Since LiDAR intensity and ray penetration behavior strongly depend on the viewing direction, spherical harmonic (SH) coefficients are employed to explicitly model the joint parameters $(\zeta,\beta_{drop},\beta_{hit})$ in an anisotropic manner.

Dynamic objects are identified via tracked bounding boxes and masked to isolate the static background. The remaining point clouds are aggregated across frames and voxel-downsampled to initialize the static model $\mathcal{G}_{bg}$.

Rigid dynamic instances such as vehicles are parameterized in an object-centric canonical space as $\mathcal{G}_{i}^{rg}$ for the $i$-th model, maintaining temporal invariance of primitive geometry. At timestamp $t$, a rigid transformation defined by rotation $R_i^t \in SO(3)$ and translation dit $\mathbf{d}_i^t\in\mathbb{R}^3$ maps the canonical mean position $\mathbf{x}_c$ and covariance $\Sigma_c$ to world coordinates as:

\begin{equation}
\label{deqn_ex2}
x_w^t=R_i^tx_c+d_i^t
\end{equation}
\begin{equation}
\label{deqn_ex3}
\Sigma_w^t=R_i^t\Sigma_c(R_i^t)^T
\end{equation}

Non-rigid entities such as pedestrians and cyclists are essential for comprehensive driving simulation yet have frequently been neglected in prior work. To faithfully represent these complex underlying geometries, non-rigid objects are decoupled into two distinct modeling paradigms, as illustrated in Fig. \ref{fig:overview}. The first is an SMPL-based joint-driven model $\mathcal{G}_{j}^{smpl}$ tailored for pedestrians, enabling fine-grained, skeleton-level control over walking and running motions. The second is a lightweight deformation model $\mathcal{G}_{k}^{dfm}$ addressing the remaining dynamic entities that lack explicit topological templates (e.g., cyclists).

\textit{1) SMPL Human Modeling:} For pedestrian instances, the SMPL model \cite{smpl} is introduced to parameterize human motion and surface deformations. The associated 2D Gaussian primitives are jointly driven by the shape parameter $\beta$ and the time-evolving pose parameter $\theta(t)\in\mathbb{R}^{24\times3\times3}$, producing a compact and efficient representation. Specifically, for the $j$-th pedestrian instance, we first instantiate the template mesh in its canonical rest pose and bind the initial set of Gaussians $\mathcal{G}_{j}^{can}$ to its surface. Subsequently, the world-space configuration of these Gaussians at timestamp $t$ can be derived as follows:
\begin{equation}
\label{deqn_ex4}
\mathcal{G}_j^{\mathrm{smpl}}(\mathfrak{t})=\mathcal{H}_j(\mathfrak{t})\otimes\mathcal{F}_{\mathcal{LBS}}(\theta(\mathfrak{t}),\mathcal{G}_j^{can})
\end{equation}
\noindent where $\mathcal{H}_j(t) \in \mathbb{SE}(3)$ denotes the global pose of this dynamic instance at time $t$, and $\mathcal{F}_{\mathcal{L}\mathcal{B}\mathcal{S}}(\cdot)$ represents the Linear Blend Skinning (LBS) operator. A skinning weight mapping is initially established between the canonical-space Gaussian primitives and the skeletal joints. These skinning weights are optimized during training via a voxel deformer \cite{avatar}, which introduces a learnable weight residual to compensate for deviations between the template skinning prior and the actual observations. This significantly enhances the non-rigid adaptability of the dynamic human Gaussian representation. The joint transformation matrices are dynamically updated according to the time-varying pose parameters $\theta$, thereby transforming the corresponding Gaussians via the optimized skinning weights. Ultimately, this mechanism continuously drives the mean positions and rotation matrices of the Gaussians, ensuring they accurately align with the human instance's pose at timestamp $t$.

Given the inherent sparsity of LiDAR point clouds, directly optimizing the high-degree-of-freedom SMPL pose parameters $\theta(t)$ from scratch is highly ill-posed. To mitigate this optimization challenge, we follow the methodology of OmniRe \cite{omnire} to derive reliable initial kinematic parameters from image-domain observations.

\textit{2) Deformable Modeling:} A general deformable representation model is introduced to establish a unified framework for multi-class non-rigid dynamic objects that are not covered by the SMPL prior. Such entities include distant pedestrians with insufficient pose cues as well as cyclists and other out-of-distribution categories lacking explicit topological templates. By characterizing local non-rigid variations in the spatiotemporal domain through a data-driven implicit deformation learning mechanism, the proposed model can represent non-rigid categories in dynamic driving scenes without relying on class-specific parametric templates.

Specifically, the general deformation network $\mathcal{D}_{\omega}$ parameterized by $\omega$ is constructed to learn time-varying local deformations of Gaussian primitives within each deformable model. For the $k$-th dynamic deformable instance at a given time $t$, its Gaussian model in the world coordinate system is defined as follows:

\begin{equation}
\label{deqn_ex5}
\mathcal{G}_k^{dfm}(t)=H_k(t)\otimes\left(\mathcal{G}_k^{can}\oplus D_\mathrm{\omega}(\mathcal{G}_k^{can},z_k,t)\right)
\end{equation}
\noindent where the deformation network $\mathcal{D}_{\omega}$ is responsible for predicting the residual offsets of attributes from the canonical-space Gaussian primitives $\mathcal{G}_k^{can}$ at timestamp $t$. Its outputs primarily consist of the positional translation $\Delta\mu_k(t)$, the rotation quaternion increment $\Delta q_k(t)$, and the spatial scale perturbation $\Delta S_k(t)$. These predicted variations are subsequently integrated with the initial $\mathcal{G}_k^{can}$ via a feature fusion operator $\oplus$. This operator performs element-wise addition on the corresponding attributes, yielding the deformed Gaussian attribute set $(\mu+\Delta\mu_k(t), q+\Delta q_k(t), S+\Delta S_k(t))$. Within this architecture, our objective is to construct a highly expressive, independent deformation field for each deformable instance. Concurrently, to ensure framework-level computational efficiency, the network parameters $\omega$ are globally shared across all deformable instances. Unique kinematic characteristics and identity disambiguation among distinct instances are resolved by incorporating a learnable instance-level embedding vector $z_k$.

\subsection{Regulated Gaussian Densification}

\begin{figure*}[t]
    \centering
    \includegraphics[width=\textwidth]{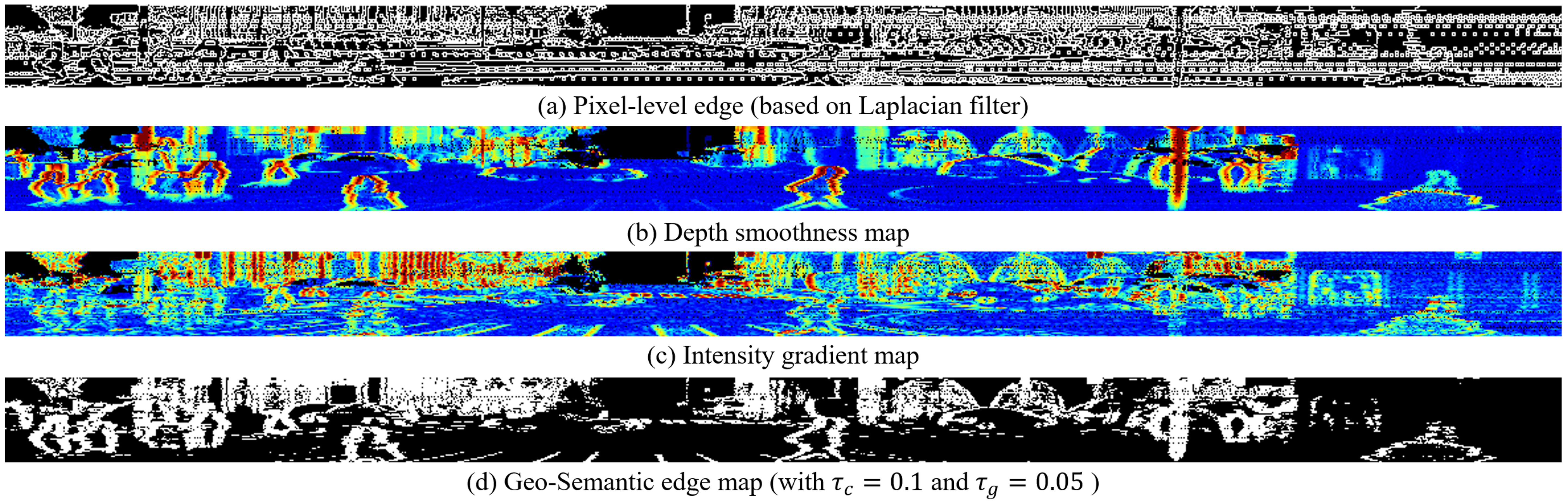}
    \caption{Comparison of LiDAR edge extraction strategies. (a) Conventional Laplacian edges. (b) Geometric depth smoothness. (c) Intensity gradient. (d) Proposed geo-semantic joint edge map ($\tau_c=0.1$, $\tau_g=0.05$), which fuses both cues to suppress clutter and isolate structurally meaningful boundaries.}
    \label{fig:edge compare}
\end{figure*}

The Adaptive Density Control (ADC) in 3DGS faces a critical dilemma in LiDAR simulation: its unconstrained growth leads to massive primitive proliferation and VRAM overflow, whereas enforcing a hard capacity limit causes premature saturation that halts densification and degrades reconstruction quality.

To address these limitations, a regulated Gaussian densification (RGD) strategy is proposed. A maximum growth budget is allocated for each step based on user-defined constraints to ensure a strictly predictable densification process. The primitive count scales with the number of densification steps according to a quadratic growth trend \cite{taming}:
\begin{equation}
\label{deqn_ex11}
Q(x) = \mathcal{B} - (\mathcal{B} - S)(1 - x/N)^2
\end{equation}
\noindent where $\mathcal{B}$ denotes the final user-defined maximum number of Gaussians (the budget), $S$ represents the initialized total number of Gaussians, and $N$ indicates the total number of densification rounds.

To guide Gaussian densification and pruning within this budgeted framework, we develop a comprehensive importance scoring mechanism. Beyond simple gradient-based heuristics, we compute a saliency score for each primitive by integrating temporal reconstruction errors, geometric attributes, visibility, and edge responses. Consequently, model capacity is preferentially allocated to regions exhibiting structural complexity, high visibility, and challenging reconstruction conditions.

To emphasize geometric discontinuities, object boundaries, and high-error regions, an edge-aware importance map $M_t \in \mathbb{R}^{H \times W}$ is computed across temporal frames. This map is constructed from two components. The first component is a pixel-level error map $M_{erro}$ derived from the discrepancy between rendered and ground-truth depth and intensity maps. The second component is an edge prior map $M_{edge}$ extracted from the LiDAR ground-truth. Departing from conventional visual edge detection methods, we introduce an edge-awareness mechanism that leverages the geometric and semantic characteristics of LiDAR. Geometric edges are initially identified based on depth smoothness. For a central pixel at coordinates $(i,j)$, the local depth smoothness $C_{i,j}$ is formulated as follows:
\begin{equation}
\label{deqn_ex12}
C_{i,j}=\frac{1}{\left|K_{i,j}\right|\cdot R_{i,j}}\left\|\sum_{k\in K_{i,j}}\left(P_{i,j}-P_{i,k}\right)\right\|
\end{equation}
\noindent where $K_{i,j}$ denotes the horizontal $K$-neighborhood, $R_{i,j}$ represents the Euclidean distance to the sensor center, and $P_{i,j}$ indicates the spatial coordinates. Based on the depth smoothness $C_{i,j}$, the image domain is partitioned into planar and edge regions. To extract additional edge regions, semantic information carried by LiDAR intensity is exploited. Considering the left $L_M$ and right $R_N$ horizontal neighborhoods of pixel $(i, j)$, the intensity gradient is computed as:
\begin{equation}
\label{deqn_ex13}
G_{i,j}=\left|\frac{1}{|\mathcal{L}_M|}\sum_{j\in\mathcal{L}_M}I_m-\frac{1}{|\mathcal{R}_N|}\sum_{k\in\mathcal{R}_N}I_N\right|
\end{equation}
\noindent where $I$ denotes the intensity value, $M$ and $N$ represent the number of valid points within the respective neighborhoods. By referencing the depth smoothness $C_{i,j}$ and the intensity gradient $G_{i,j}$, LiDAR edge regions are jointly determined using predefined segmentation thresholds $\tau_c$ and $\tau_g$. This detection serves as a critical component in calculating the Gaussian saliency score. As illustrated in Fig. \ref{fig:edge compare}, the proposed geo-semantic edge map captures LiDAR-specific discontinuities and clutter more accurately than conventional pixel-level edge maps.

Let $\mathcal{G}=\{g_i \}_{i=1}^N$ denote a scene containing $N$ Gaussian primitives. The overall importance score for each primitive $g_i$ is defined as:
\begin{equation}
\label{deqn_ex14}
S_i=\frac{1}{|T_i|}\sum_{t\in T_i}\mathcal{E}_t(S_i^{attr}+S_{t,i}^{ray})
\end{equation}
\noindent where $T_i \subseteq T$ denotes the subset of scene frames in which Gaussian $g_i$ contributes to LiDAR ray rendering. This subset mitigates the bias arising from varying observation frequencies of Gaussians across temporal frames. $\mathcal{E}_t$ represents the aggregate LiDAR reconstruction error for the $t$-th frame, calculated via a combination of $L_1$ and SSIM losses on depth and intensity maps. $S_i^{attr}$ indicates the attribute importance of the Gaussian primitive, derived from a weighted combination of the primitive's positional gradient $\nabla_i$, opacity $o_i$, and scale $S_i$. $S_{t,i}^{ray}$ denotes the importance of the rendering contribution along each ray at timestamp $t$. This importance is propagated to each Gaussian through an auxiliary rendering process and is computed as follows:
\begin{equation}
\label{deqn_ex15}
S_{t,i}^{ray}=\gamma_w\mathcal{W}_{t,i}+\gamma_dd_{t,i}+\gamma_cc_{t,i}+\gamma_pp_{t,i}
\end{equation}
\noindent where $\mathcal{W}_{t,i}$ represents the accumulated contribution weight of Gaussian $g_i$ during LiDAR rendering at timestamp $t$, $d_{t,i}$ denotes the distance-related accumulation, $c_{t,i}$ indicates the frequency of ray-Gaussian intersections, $p_{t,i}$ denotes the accumulated blending weight of the image-level feature map $M_t$ at the location of Gaussian $g_i$, and $\gamma$ serves as a correlation coefficient. 

The final score $S_i$ of each Gaussian primitive is converted into a sampling probability distribution. A weighted random resampling strategy is then employed to sample the target number of candidate primitives from the existing model for adaptive scene densification.

\subsection{Optimization}
To align the 2DGS distribution with LiDAR physical properties, all learnable parameters within the dynamic scene graph, including foundational Gaussian attributes and kinematic variables for non-rigid modeling, are optimized end-to-end via the following joint loss:
\begin{equation}
\begin{aligned}
\label{deqn_ex16}
\mathcal{L}=\lambda_d\mathcal{L}_d^1+\lambda_{md}\mathcal{L}_{md}^1+(1-\lambda_i)\mathcal{L}_i^1+\lambda_i\mathcal{L}_i^{SSIM}\\+\lambda_{dt}\mathcal{L}_{dt}^{avg}+\lambda_{rd}\mathcal{L}_{rd}^{BCE}+\lambda_{pc}\mathcal{L}_{pc}^{CD}+\lambda_{reg}\mathcal{L}_{reg}
\end{aligned}
\end{equation}
\noindent where $\lambda_*$ are the corresponding weight coefficients. For $\mathcal{L}_*^*$, the subscript and superscript denote the loss type and computation method, respectively. $\mathcal{L}_{reg}$ aggregates dynamic object regularizations, including out-of-box, smoothness, and neighborhood penalties. To enhance the compactness and physical fidelity of reconstructed surfaces, we apply two geometric constraints: a median depth loss $\mathcal{L}_{md}^1$ \cite{gslidar} to concentrate ray density on physical boundaries, and a 2DGS distortion loss $\mathcal{L}_{dt}^{avg}$ \cite{2dgs} to penalize excessive ray intersection distances.

Following existing pipelines \cite{gslidar,LiDAR-RT}, a U-Net is employed post-training to refine the simulated ray-drop effect. Taking the rendered probability, depth, and intensity maps as input, it predicts the final ray-drop mask under binary cross-entropy (BCE) supervision.

\section{Experiments}
\subsection{Experimental Setup}

URL-Sim is validated on the widely used Waymo Open Dataset \cite{waymo} and KITTI-360 \cite{kitti}. The Waymo data were collected at 10 Hz using a 64-beam LiDAR with a 20° vertical field of view (FOV), natively providing depth and intensity images at a resolution of 64×2560. KITTI-360 employs a 64-beam LiDAR operating at 10 Hz with a 26.4° vertical FOV. The raw KITTI-360 LiDAR data are spherically projected into 2D depth and intensity images at a 64×1030 resolution. Four dynamic scene sequences containing non-rigid dynamic instances were selected from each dataset. For each sequence, 50 consecutive frames were extracted, with every tenth frame allocated to the test set and the remainder used for training.

Comprehensive model evaluation relies on a quantitative framework that targets two primary dimensions: 2D perceptual quality and 3D geometric accuracy. For 2D projections, pixel-level depth and intensity deviations are measured via Root Mean Square Error (RMSE) and Median Absolute Error (MedAE), with global fidelity assessed through SSIM \cite{ssim}, Peak Signal-to-Noise Ratio (PSNR), and LPIPS \cite{lpips}. In 3D space, structural discrepancy is measured using Chamfer Distance (CD) \cite{cd} and F-Score at a 5 cm threshold. Rendering efficiency is evaluated via frame rate (FPS) and total Gaussian primitive count.

As the proposed RGD is seamlessly integrated into the training process, the densification setup aligns completely with the original 2DGS. Note that SMPL-based human models, constrained by a fixed topological template, are excluded from the densification process. To respect memory constraints, the primitive count is capped at 3 million Gaussians, a budget that is also applied to our proposed RGD strategy to ensure a fair comparison.

\begin{table*}[t]
\centering
% 标题中增加了高亮图例，使用 \protect 防止在 caption 中报错
\caption{Ablation studies on dynamic scene of the Waymo Open Dataset \cite{waymo}. Cells are highlighted as follows: \protect\colorbox{bestcolor}{BEST} and \protect\colorbox{secondcolor}{SECOND BEST}.“GN” in the table stands for number of Gaussian primitives}
\label{tab:1}
\resizebox{\textwidth}{!}{
\begin{tabular}{l cc ccccc ccccc cc}
\toprule
\multirow{2}{*}{Method} & \multicolumn{2}{c}{Efficiency} & \multicolumn{5}{c}{Depth} & \multicolumn{5}{c}{Intensity} & \multicolumn{2}{c}{Point Cloud} \\
\cmidrule(lr){2-3} \cmidrule(lr){4-8} \cmidrule(lr){9-13} \cmidrule(lr){14-15}
& FPS$\uparrow$ & GN$\downarrow$ & RMSE $\downarrow$ & MedAE $\downarrow$ & LPIPS $\downarrow$ & SSIM $\uparrow$ & PSNR $\uparrow$ & RMSE $\downarrow$ & MedAE $\downarrow$ & LPIPS $\downarrow$ & SSIM $\uparrow$ & PSNR $\uparrow$ & CD $\downarrow$ & F-score $\uparrow$ \\
\midrule
Rigid & \cellcolor{bestcolor} 20.1 & 3.17M & 6.3597 & 0.0220 & 0.2697 & 0.7004 & 22.0404 & 0.0586 & 0.0061 & 0.1912 & 0.7827 & 25.0785 & 0.0863 & 0.9461 \\
w/o NRM & 19.7 & 3.00M & \cellcolor{secondcolor} 6.1119 & \cellcolor{secondcolor} 0.0187 & \cellcolor{secondcolor} 0.2564 & 0.7079 & \cellcolor{secondcolor} 22.3780 & \cellcolor{secondcolor} 0.0546 & \cellcolor{secondcolor} 0.0055 & \cellcolor{secondcolor} 0.1513 & \cellcolor{secondcolor} 0.8071 & \cellcolor{secondcolor} 26.0181 & \cellcolor{secondcolor} 0.0710 & 0.9543 \\
w/o RGD & \cellcolor{secondcolor} 18.2 & \cellcolor{secondcolor} 3.09M & 6.2186 & 0.0238 & 0.2630 & \cellcolor{secondcolor} 0.7081 & 22.2407 & 0.0551 & 0.0058 & 0.1723 & 0.7861 & 25.6559 & 0.0717 & \cellcolor{secondcolor} 0.9562 \\
\makecell[l]{URL-Sim} & 17.6 & \cellcolor{bestcolor} 3.00M & \cellcolor{bestcolor} 5.9844 & \cellcolor{bestcolor} 0.0184 & \cellcolor{bestcolor} 0.2042 & \cellcolor{bestcolor} 0.7154 & \cellcolor{bestcolor} 22.5603 & \cellcolor{bestcolor} 0.0503 & \cellcolor{bestcolor} 0.0052 & \cellcolor{bestcolor} 0.1377 & \cellcolor{bestcolor} 0.8259 & \cellcolor{bestcolor} 26.5183 & \cellcolor{bestcolor} 0.0468 & \cellcolor{bestcolor} 0.9682 \\
\bottomrule
\end{tabular}
}
\end{table*}

\begin{table*}[t]
\centering
\caption{Ablation studies on dynamic scene of the KITTI-360 Dataset \cite{kitti}}
\label{tab:2}
% 使用 resizebox 将表格宽度限制在文本宽度内，防止溢出
\resizebox{\textwidth}{!}{
\begin{tabular}{l cc ccccc ccccc cc}
\toprule
\multirow{2}{*}{Method} & \multicolumn{2}{c}{Efficiency} & \multicolumn{5}{c}{Depth} & \multicolumn{5}{c}{Intensity} & \multicolumn{2}{c}{Point Cloud} \\
\cmidrule(lr){2-3} \cmidrule(lr){4-8} \cmidrule(lr){9-13} \cmidrule(lr){14-15}
& FPS$\uparrow$ & GN$\downarrow$ & RMSE $\downarrow$ & MedAE $\downarrow$ & LPIPS $\downarrow$ & SSIM $\uparrow$ & PSNR $\uparrow$ & RMSE $\downarrow$ & MedAE $\downarrow$ & LPIPS $\downarrow$ & SSIM $\uparrow$ & PSNR $\uparrow$ & CD $\downarrow$ & F-score $\uparrow$ \\
\midrule
Rigid & \cellcolor{bestcolor} 24.6 & 3.06M & 4.0220 & 0.0341 & 0.0749 & 0.9010 & 26.1136 & 0.0940 & 0.0284 & 0.1879 & 0.6751 & 20.5697 & 0.1367 & 0.9201 \\
w/o NRM & 24.1 & 3.00M & \cellcolor{secondcolor} 3.4800 & \cellcolor{bestcolor} 0.0276 & \cellcolor{secondcolor} 0.0641 & \cellcolor{secondcolor} 0.9156 & \cellcolor{secondcolor} 27.3287 & \cellcolor{secondcolor} 0.0846 & \cellcolor{secondcolor} 0.0235 & \cellcolor{secondcolor} 0.1420 & \cellcolor{bestcolor} 0.7347 & \cellcolor{secondcolor} 21.5308 & \cellcolor{secondcolor} 0.0942 & \cellcolor{secondcolor} 0.9441 \\
w/o RGD & 22.7 & \cellcolor{secondcolor} 3.12M & 3.8669 & 0.0340 & 0.0708 & 0.9041 & 26.3839 & 0.0915 & 0.0270 & 0.1766 & 0.6912 & 20.8249 & 0.1340 & 0.9235 \\
\makecell[l]{URL-Sim} & \cellcolor{secondcolor} 22.3 & \cellcolor{bestcolor} 3.00M & \cellcolor{bestcolor} 3.4700 & \cellcolor{secondcolor} 0.0277 & \cellcolor{bestcolor} 0.0583 & \cellcolor{bestcolor} 0.9162 & \cellcolor{bestcolor} 27.3792 & \cellcolor{bestcolor} 0.0844 & \cellcolor{bestcolor} 0.0233 & \cellcolor{bestcolor} 0.1404 & \cellcolor{secondcolor} 0.7334 & \cellcolor{bestcolor} 21.5475 & \cellcolor{bestcolor} 0.0909 & \cellcolor{bestcolor} 0.9447 \\
\bottomrule
\end{tabular}
}
\end{table*}

\subsection{Ablation Studies}

\begin{figure*}[t]
    \centering
    \includegraphics[width=\textwidth]{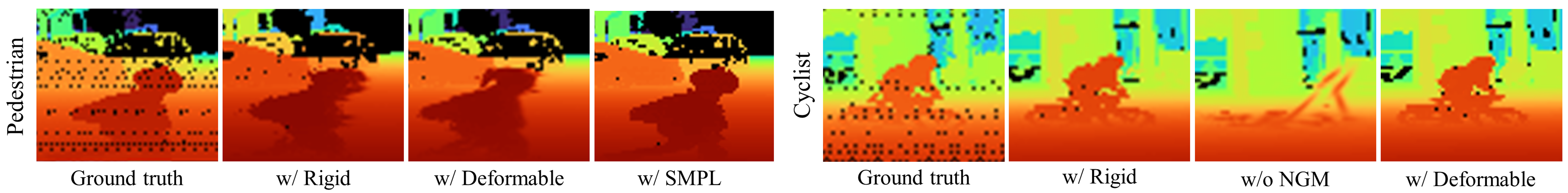}
    \caption{Ablation of non-rigid modeling.}
    \label{fig:nrm}
\end{figure*}

\begin{figure}[!t] 
    \centering
    \includegraphics[width=\columnwidth]{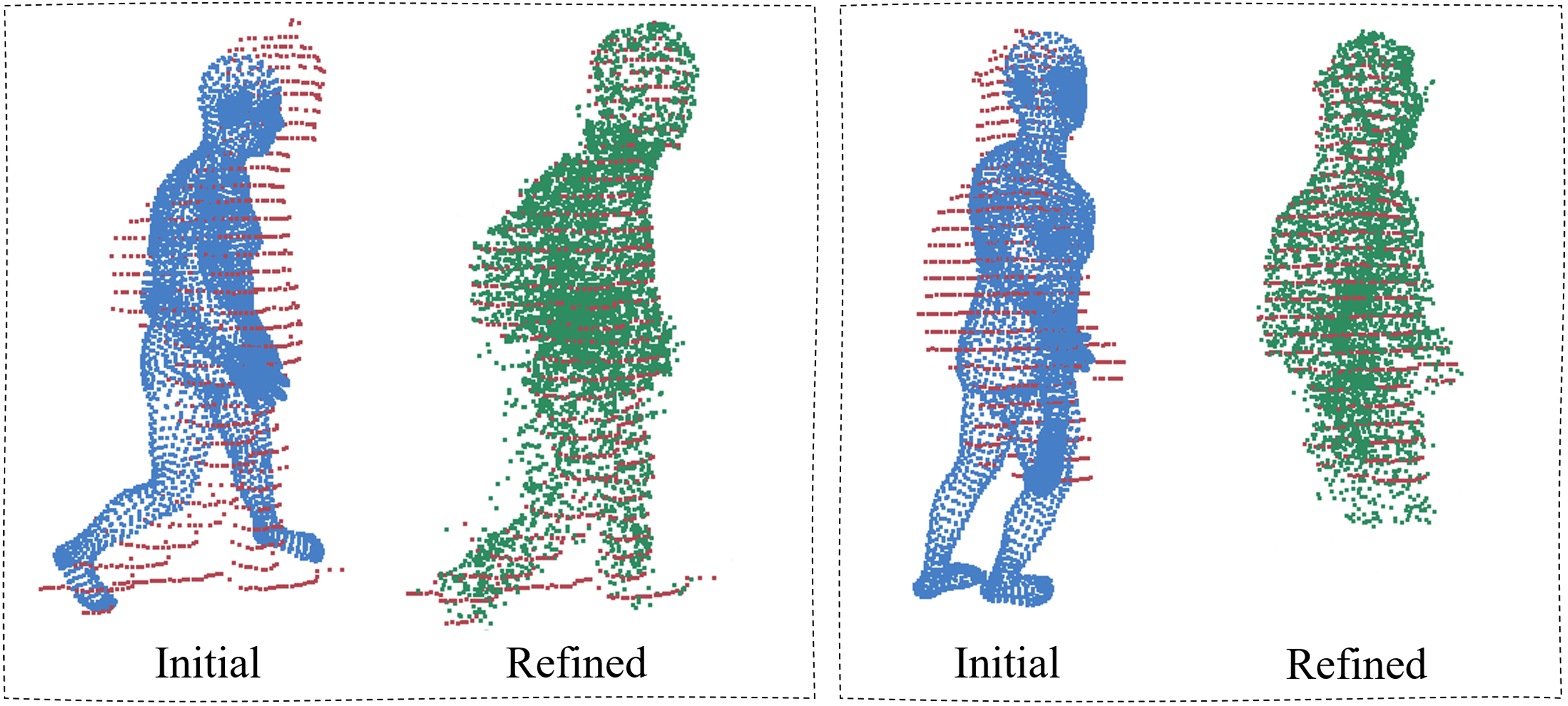}
    \caption{Ablation of SMPL human refinement. The \textcolor{red}{red} point clouds represent the ground truth LiDAR point clouds of human instances, while the \textcolor{blue}{blue} and \textcolor{green}{green} point clouds denote the Gaussian means of the human models upon SMPL initialization and after optimization, respectively.}
    \label{fig:smpl}
\end{figure}

\begin{table}[t]
\centering
\caption{Ablation studies on different components of RGD. "RS" in the table stands for random sampling based on ADC}
\label{tab:ablation_rgd}
\resizebox{\linewidth}{!}{
\begin{tabular}{l ccc ccc cc}
\toprule
\multirow{2}{*}{Method} & \multicolumn{3}{c}{Depth} & \multicolumn{3}{c}{Intensity} & \multicolumn{2}{c}{Point Cloud} \\
\cmidrule(lr){2-4} \cmidrule(lr){5-7} \cmidrule(lr){8-9}
& RMSE $\downarrow$ & SSIM $\uparrow$ & PSNR $\uparrow$ & RMSE $\downarrow$ & SSIM $\uparrow$ & PSNR $\uparrow$ & CD $\downarrow$ & F-score $\uparrow$ \\
\midrule
ADC & 4.1685 & 0.8716 & 25.6781 & 0.1062 & 0.6614 & 19.4904 & 0.1243 & 0.9388 \\
RS & 4.1599 & 0.8700 & 25.6994 & 0.1076 & 0.6463 & 19.3746 & 0.1326 & 0.9352 \\
w/o $S^{ray}$ & 3.9513 & 0.8827 & 26.1966 & 0.0993 & 0.7005 & 19.9882 & 0.0871 & 0.9566 \\
w/o edge & 3.9102 & 0.8859 & 26.2679 & 0.0990 & 0.7085 & 20.0121 & 0.0858 & 0.9573 \\
w/o opacity & 3.9366 & 0.8826 & 26.2003 & 0.1013 & 0.7062 & 20.0309 & 0.0875 & 0.9566 \\
w/o scale & \cellcolor{secondcolor} 3.8207 & 0.8885 & \cellcolor{secondcolor} 26.4212 & \cellcolor{secondcolor} 0.0986 & \cellcolor{secondcolor} 0.7099 & \cellcolor{secondcolor} 20.1541 & 0.0868 & 0.9567 \\
w/o grad & 3.8430 & \cellcolor{bestcolor} 0.8898 & 26.4124 & 0.0998 & 0.7091 & 20.1323 & \cellcolor{secondcolor} 0.0816 & \cellcolor{secondcolor} 0.9574 \\
RGD & \cellcolor{bestcolor} 3.7871 & \cellcolor{secondcolor} 0.8894 & \cellcolor{bestcolor} 26.5406 & \cellcolor{bestcolor} 0.0983 & \cellcolor{bestcolor} 0.7140 & \cellcolor{bestcolor} 20.1878 & \cellcolor{bestcolor} 0.0798 & \cellcolor{bestcolor} 0.9577 \\
\bottomrule
\end{tabular}
}
\end{table}

Ablation studies are conducted under identical configurations to evaluate Non-Rigid Modeling (NRM) and the RGD strategy. The baseline, denoted as "Rigid", consists solely of the static background and rigid vehicles with conventional Adaptive Density Control (ADC). Quantitative results in Table \ref{tab:1} and \ref{tab:2} demonstrate that explicit non-rigid modeling facilitates faithful reconstruction of dynamic entities and yields significant improvements in perceptual quality. The RGD strategy restricts the total primitive count to 3 million while achieving superior performance and parameter efficiency over standard ADC.

\textbf{Non-rigid modeling.} Fig. \ref{fig:nrm} presents a qualitative comparison of different modeling strategies. Applying rigid modeling to non-rigid objects induces severe geometric ghosting and structural collapse. For pedestrians with complex articulated motions, the SMPL-based joint-driven model provides robust topological constraints, enabling precise restoration of real contours and dynamic poses. For template-free objects such as cyclists, the deformation network captures fine-scale local deformations, eliminating geometric distortions that arise when such entities are processed as rigid bodies.

Fig. \ref{fig:smpl} reveals clear pose errors and spatial misalignments between initial image-derived human poses and ground-truth LiDAR point clouds. Since these initial estimates may degrade reconstruction quality, the pose and shape parameters are jointly refined. The optimized point cloud achieves closer alignment with the ground truth and better preserves the actual human geometry and pose.

\textbf{Regulated Gaussian densification}. Table \ref{tab:ablation_rgd} details the ablation studies for the RGD strategy. The baseline ADC yields suboptimal performance across depth and intensity metrics. Directly applying random sampling (RS) to constrain the budget avoids severe metric degradation but fails to enhance reconstruction quality. The full RGD strategy achieves optimal performance across all metrics. Removing individual guidance components causes varying degrees of performance degradation. These results indicate that RGD effectively directs the computational budget toward regions with geometric discontinuities and physical boundaries, improving accuracy and fidelity under a strictly bounded primitive count.

\begin{table*}[t]
\centering
\caption{Quantitative comparison with SOTAs on the Waymo Open Dataset \cite{waymo}. Metrics are averaged over selected four dynamic scenes.}
\label{tab:sota_waymo}
% 使用 resizebox 将表格宽度限制在文本宽度内，防止溢出
\resizebox{\textwidth}{!}{
\begin{tabular}{l cc ccccc ccccc cc}
\toprule
\multirow{2}{*}{Method} & \multicolumn{2}{c}{Efficiency} & \multicolumn{5}{c}{Depth} & \multicolumn{5}{c}{Intensity} & \multicolumn{2}{c}{Point Cloud} \\
\cmidrule(lr){2-3} \cmidrule(lr){4-8} \cmidrule(lr){9-13} \cmidrule(lr){14-15}
& FPS $\uparrow$ & GN $\downarrow$ & RMSE $\downarrow$ & MedAE $\downarrow$ & LPIPS $\downarrow$ & SSIM $\uparrow$ & PSNR $\uparrow$ & RMSE $\downarrow$ & MedAE $\downarrow$ & LPIPS $\downarrow$ & SSIM $\uparrow$ & PSNR $\uparrow$ & CD $\downarrow$ & F-score $\uparrow$ \\
\midrule
DyNFL\cite{DyNFL} & 0.2 & - & 7.6841 & 0.0319 & 0.2764 & 0.6673 & 20.5682 & 0.0613 & 0.0066 & 0.1586 & 0.7538 & 24.6592 & 0.2037 & 0.6441 \\
LiDAR-Nerf\cite{LiDAR-NeRF} & 1.7 & - & 9.2613 & 0.2529 & 0.4283 & 0.5669 & 19.4217 & 0.0748 & 0.0169 & 0.2466 & 0.6349 & 22.7140 & 1.6044 & 0.7351 \\
LiDAR4D\cite{LiDAR4D} & 0.2 & - & \cellcolor{secondcolor} 6.2431 & 0.0262 & \cellcolor{bestcolor} 0.1694 & \cellcolor{secondcolor} 0.7018 & \cellcolor{secondcolor} 21.8803 & \cellcolor{secondcolor} 0.0536 & 0.0084 & \cellcolor{secondcolor} 0.1500 & \cellcolor{secondcolor} 0.7927 & \cellcolor{secondcolor} 25.2151 & 0.0886 & 0.9201 \\
LiDAR-GS\cite{lidargs} & 9.7 & \cellcolor{secondcolor} 3.13M & 9.4453 & 0.2082 & 0.4132 & 0.6141 & 19.1070 & 0.0752 & 0.0113 & 0.2362 & 0.6543 & 22.6243 & 1.3034 & 0.7505 \\
GS-LiDAR\cite{gslidar} & 13.5 & 3.25M & 6.5128 & 0.0622 & 0.2575 & 0.6982 & 21.8211 & 0.0663 & 0.0079 & 0.2363 & 0.7089 & 23.7152 & 0.4277 & 0.7986 \\
LiDAR-RT\cite{LiDAR-RT} & \cellcolor{bestcolor} 20.1 & 3.17M & 6.3597 & \cellcolor{secondcolor} 0.0220 & 0.2697 & 0.7004 & 22.0404 & 0.0586 & \cellcolor{secondcolor} 0.0061 & 0.1912 & 0.7827 & 25.0785 & \cellcolor{secondcolor} 0.0863 & \cellcolor{secondcolor} 0.9461 \\
\textbf{URL-Sim (ours)} & \cellcolor{secondcolor} 17.6 & \cellcolor{bestcolor} 3.00M & \cellcolor{bestcolor} 5.9844 & \cellcolor{bestcolor} 0.0184 & \cellcolor{secondcolor} 0.2042 & \cellcolor{bestcolor} 0.7154 & \cellcolor{bestcolor} 22.5603 & \cellcolor{bestcolor} 0.0503 & \cellcolor{bestcolor} 0.0052 & \cellcolor{bestcolor} 0.1377 & \cellcolor{bestcolor} 0.8259 & \cellcolor{bestcolor} 26.5183 & \cellcolor{bestcolor} 0.0468 & \cellcolor{bestcolor} 0.9682 \\
\bottomrule
\end{tabular}
}
\end{table*}

\begin{table*}[t]
\centering
\caption{Quantitative comparison with SOTAs on the KITTI-360 dataset\cite{kitti}. Metrics are averaged over selected four dynamic scenes.}
\label{tab:sota_kitti}
\resizebox{\textwidth}{!}{
\begin{tabular}{l cc ccccc ccccc cc}
\toprule
\multirow{2}{*}{Method} & \multicolumn{2}{c}{Efficiency} & \multicolumn{5}{c}{Depth} & \multicolumn{5}{c}{Intensity} & \multicolumn{2}{c}{Point Cloud} \\
\cmidrule(lr){2-3} \cmidrule(lr){4-8} \cmidrule(lr){9-13} \cmidrule(lr){14-15}
& FPS $\uparrow$ & GN $\downarrow$ & RMSE $\downarrow$ & MedAE $\downarrow$ & LPIPS $\downarrow$ & SSIM $\uparrow$ & PSNR $\uparrow$ & RMSE $\downarrow$ & MedAE $\downarrow$ & LPIPS $\downarrow$ & SSIM $\uparrow$ & PSNR $\uparrow$ & CD $\downarrow$ & F-score $\uparrow$ \\
\midrule
DyNFL\cite{DyNFL} & 0.4 & - & 6.5314 & 0.1038 & 0.2588 & 0.6557 & 21.9115 & 0.1149 & 0.0428 & 0.2859 & 0.3573 & 18.8066 & 0.3416 & 0.4776 \\
LiDAR-Nerf\cite{LiDAR-NeRF} & 5.3 & - & 4.8113 & 0.0917 & 0.1207 & 0.7429 & 24.4380 & 0.1169 & 0.0401 & 0.2397 & 0.5062 & 18.6800 & 0.2171 & 0.8314 \\
LiDAR4D\cite{LiDAR4D} & 0.4 & - & 4.1722 & 0.0386 & 0.0780 & 0.8762 & 25.6880 & 0.0952 & 0.0306 & \cellcolor{secondcolor} 0.1585 & 0.6685 & 20.4485 & 0.1930 & 0.8826 \\
LiDAR-GS\cite{lidargs} & 10.4 & 3.26M & 7.1220 & 0.0903 & 0.3227 & 0.5397 & 21.0203 & 0.1569 & 0.0721 & 0.3076 & 0.2523 & 16.0927 & 0.2482 & 0.8432 \\
GS-LiDAR\cite{gslidar} & 13.5 & 3.19M & 4.9931 & \cellcolor{secondcolor} 0.0289 & 0.0819 & \cellcolor{secondcolor} 0.9033 & 24.1976 & \cellcolor{secondcolor} 0.0870 & \cellcolor{bestcolor} 0.0216 & 0.1973 & 0.6701 & \cellcolor{secondcolor} 21.2212 & 0.4029 & 0.8787 \\
LiDAR-RT\cite{LiDAR-RT} & \cellcolor{bestcolor} 24.6 & \cellcolor{secondcolor} 3.06M & \cellcolor{secondcolor} 4.0220 & 0.0341 & \cellcolor{secondcolor} 0.0749 & 0.9010 & \cellcolor{secondcolor} 26.1136 & 0.0940 & 0.0284 & 0.1879 & \cellcolor{secondcolor} 0.6751 & 20.5697 & \cellcolor{secondcolor} 0.1367 & \cellcolor{secondcolor} 0.9201 \\
\textbf{URL-Sim (ours)} & \cellcolor{secondcolor} 22.3 & \cellcolor{bestcolor} 3.00M & \cellcolor{bestcolor} 3.4700 & \cellcolor{bestcolor} 0.0277 & \cellcolor{bestcolor} 0.0583 & \cellcolor{bestcolor} 0.9162 & \cellcolor{bestcolor} 27.3792 & \cellcolor{bestcolor} 0.0844 & \cellcolor{secondcolor} 0.0233 & \cellcolor{bestcolor} 0.1404 & \cellcolor{bestcolor} 0.7334 & \cellcolor{bestcolor} 21.5475 & \cellcolor{bestcolor} 0.0909 & \cellcolor{bestcolor} 0.9447 \\
\bottomrule
\end{tabular}
}
\end{table*}

\subsection{Comparison with the SOTAs}

\begin{figure*}[t]
    \centering
    \includegraphics[width=\textwidth]{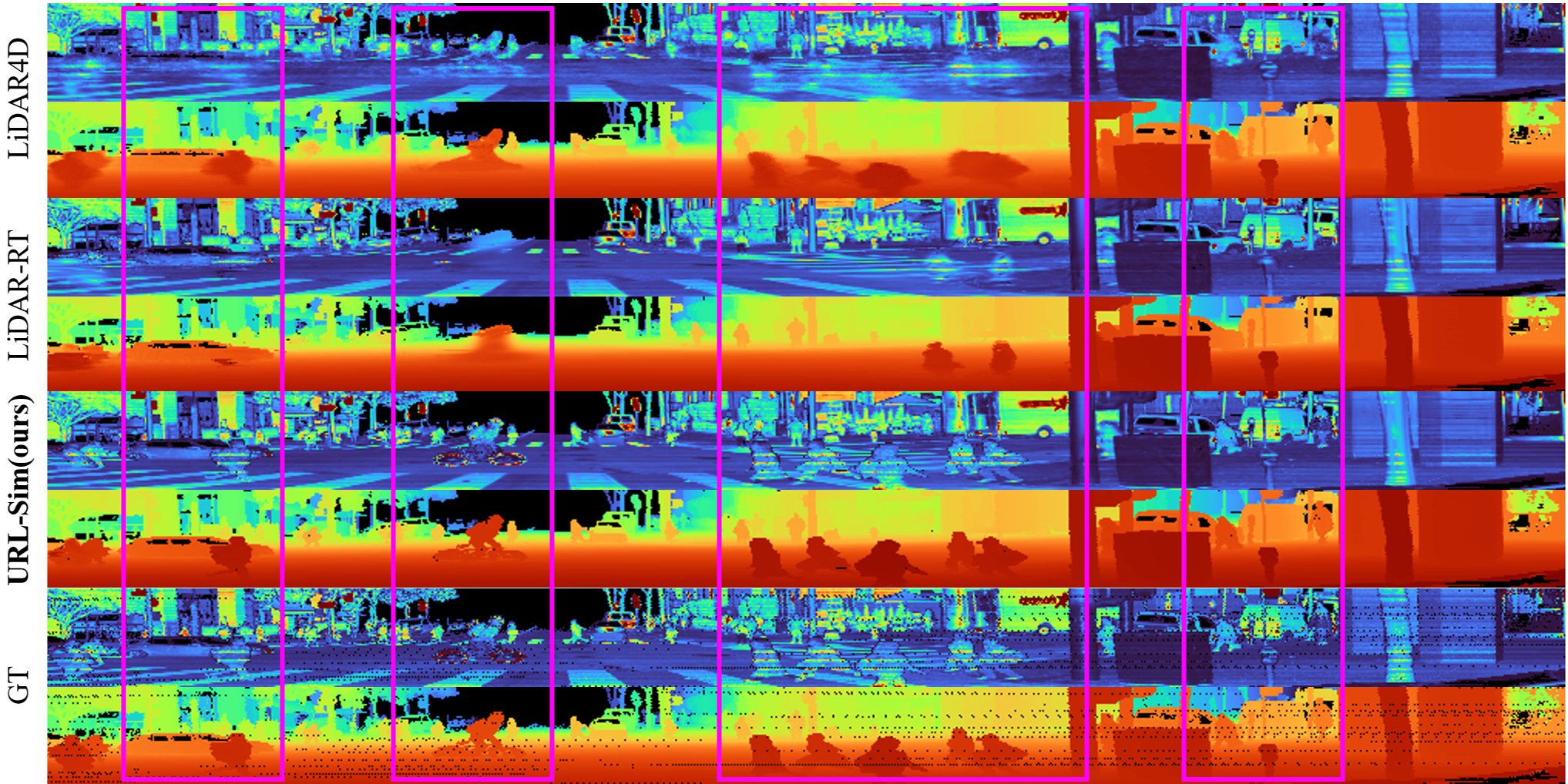}
    \caption{Qualitative comparison for LiDAR depth simulation on Waymo Open Dataset \cite{waymo} sequence seg-104859.}
    \label{fig:sota_img}
\end{figure*}

\begin{figure*}[t]
    \centering
    \includegraphics[width=\textwidth]{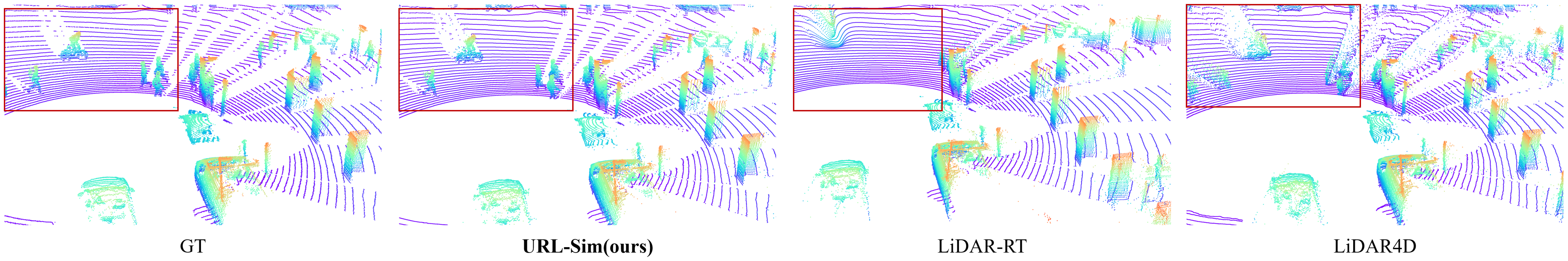}
    \caption{Qualitative comparison for LiDAR point cloud simulation on Waymo Open Dataset \cite{waymo}.}
    \label{fig:sota_pc}
\end{figure*}

URL-Sim is evaluated against state-of-the-art (SOTA) NeRF- and 3DGS-based frameworks, including LiDAR-NeRF \cite{LiDAR-NeRF}, DyNFL \cite{DyNFL}, LiDAR4D \cite{LiDAR4D}, LiDAR-GS \cite{lidargs}, GS-LiDAR \cite{gslidar}, and LiDAR-RT \cite{LiDAR-RT}. All baselines were implemented using their official open-source codebases and assessed using a unified set of quantitative metrics. Consistent pseudo-color mapping schemes are applied for visualizing depth and intensity maps. All experiments were run on a single NVIDIA RTX 5070 Ti GPU for 30,000 iterations.

Tables \ref{tab:sota_waymo} and \ref{tab:sota_kitti} summarize the quantitative results on the Waymo and KITTI-360 datasets. All 3DGS-based baselines are capped at a budget of approximately 3 million Gaussians for fair comparison. URL-Sim achieves competitive rendering speeds while processing non-rigid dynamic instances. It achieves the best performance on most 2D image rendering metrics with substantial improvements in intensity and depth PSNR. These gains are driven by effective non-rigid dynamic modeling and selective optimization of high-contribution Gaussians guided by RGD. Qualitative comparisons in Fig. \ref{fig:sota_img} demonstrate that previous methods struggle with geometric smearing and ghosting when handling non-rigid objects. URL-Sim recovers clear depth profiles and structural intensity details for non-rigid objects. The resulting reconstructions exhibit substantially sharper geometric boundaries that are consistent with the ground-truth spatial structure and physical intensity distributions.

URL-Sim also achieves the highest 3D geometric accuracy, with the lowest Chamfer Distance and highest F-score on both datasets. Compared to the second-best baseline LiDAR-RT \cite{LiDAR-RT} (0.0863 and 0.1367), URL-Sim reduces CD error by approximately 45.7\% and 33.5\% on the Waymo and KITTI-360 datasets, confirming more accurate recovery of ground-truth 3D geometry. Qualitative point cloud comparisons in Fig. \ref{fig:sota_pc} further illustrate these advantages. Previous methods either eliminate certain dynamic entities or introduce severe outliers and scanline distortions. URL-Sim demonstrates high-fidelity preservation of both LiDAR scan-line patterns and global scene topology, effectively reconstructing non-rigid actors with sharp boundaries and suppressed local noise.

\subsection{Applications of LiDAR Simulation.}

\begin{figure}[!t]
    \centering
    \includegraphics[width=\linewidth]{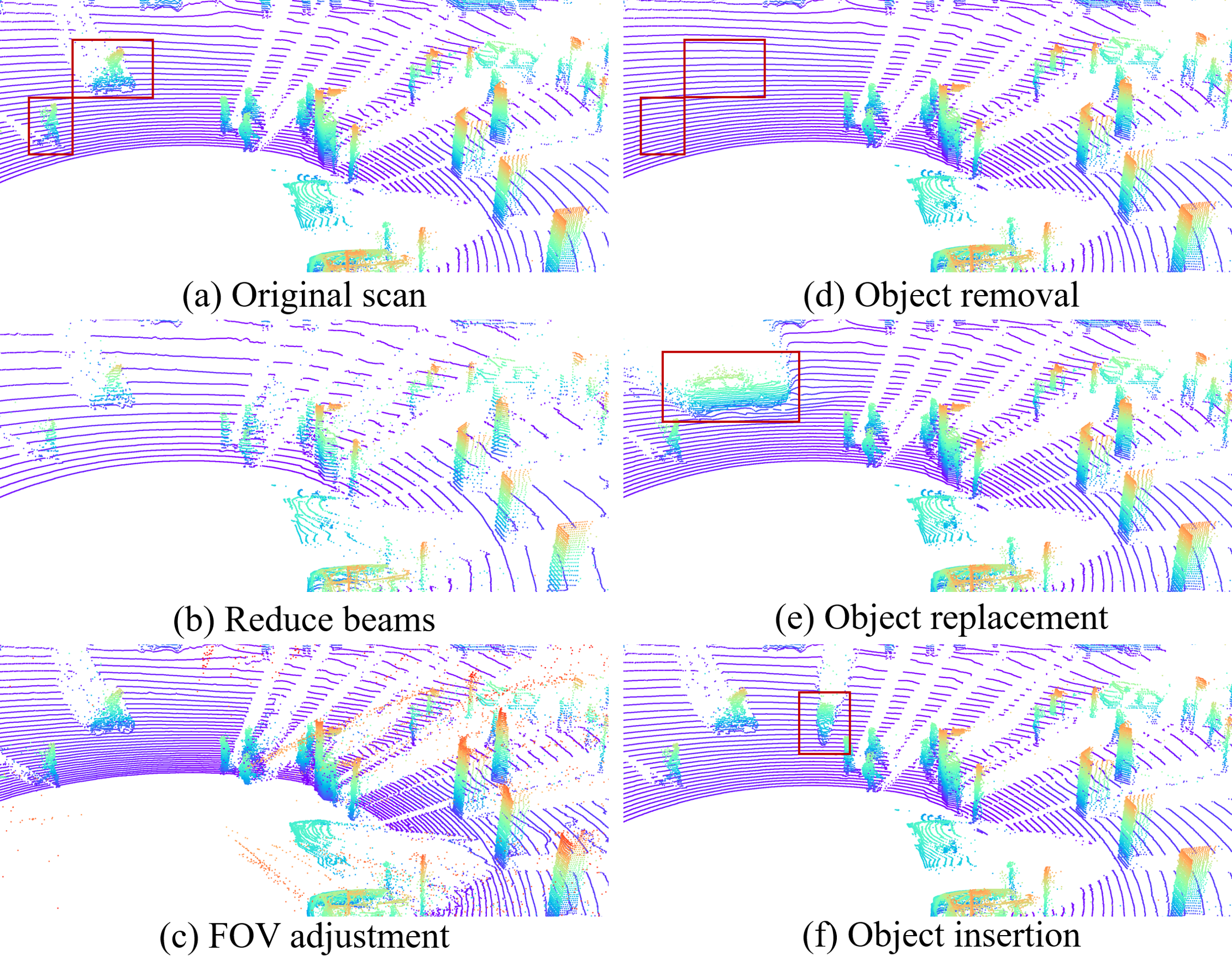}
    \caption{LiDAR re-simulation and scene editing results.}
    \label{fig:edit}
\end{figure}

Flexible LiDAR re-simulation is achieved in URL-Sim by dynamically modifying ray emission configurations to project point clouds under varying sensor specifications. Adjustments to scanning beams and vertical FOV are demonstrated in Fig. \ref{fig:edit}(b) and \ref{fig:edit}(c) to support low-cost data augmentation and cross-sensor generalization evaluations. The independent modeling of backgrounds and instances within the dynamic scene graph natively supports flexible scene editing, facilitating spatial transformations and object replacement. Specific editing operations are presented in the second column of Fig. \ref{fig:edit} to illustrate visually plausible object removal and seamless object replacement alongside novel object insertion. Together, these operations enable the synthesis of diverse interactive scenarios, providing a potential source of simulated data for future closed-loop testing and safety validation.

\section{conclusion}

This paper presents URL-Sim, a unified and regulated LiDAR simulation framework that builds a dynamic Gaussian scene graph for decoupled modeling and high-fidelity reconstruction of heterogeneous traffic participants. Integrating the SMPL prior with a lightweight deformation network bridges the gap in non-rigid LiDAR simulation, while the proposed RGD strategy employs multi-dimensional importance scoring and fixed budget
allocation to direct resources toward critical regions under a fixed primitive budget. Extensive experiments on Waymo and KITTI-360 demonstrate significant improvements over existing baselines in both perceptual quality and geometric accuracy.

Although URL-Sim delivers competitive dynamic LiDAR simulation, certain limitations remain. First, the dynamic decoupling relies on accurate bounding boxes and tracking, and reconstruction quality may degrade in crowded environments. Additionally, large viewpoint shifts during editing may distort unobserved regions. Integrating generative priors such as diffusion models represents a promising direction for overcoming these extrapolation challenges.

\bibliographystyle{IEEEtran}
\bibliography{reference}

\end{document}